\documentclass{article}

\usepackage{iclr2019_conference}
\iclrfinalcopy

\usepackage[utf8]{inputenc} 
\usepackage[T1]{fontenc}    
\usepackage{hyperref}       
\usepackage{url}            
\usepackage{booktabs}       
\usepackage{amsfonts}       
\usepackage{nicefrac}       
\usepackage{microtype}      
\usepackage{tikz}
\usepackage{pgfplotstable}
\usetikzlibrary{arrows}
\pgfplotsset{compat=1.14}
\usepgfplotslibrary{statistics}

\usetikzlibrary{decorations.markings}
\tikzset{
    set arrow inside/.code={\pgfqkeys{/tikz/arrow inside}{#1}},
    set arrow inside={end/.initial=>, opt/.initial=},
    /pgf/decoration/Mark/.style={
        mark/.expanded=at position #1 with
        {
            \noexpand\arrow[\pgfkeysvalueof{/tikz/arrow inside/opt}]{\pgfkeysvalueof{/tikz/arrow inside/end}}
        }
    },
    arrow inside/.style 2 args={
        set arrow inside={#1},
        postaction={
            decorate,decoration={
                markings,Mark/.list={#2}
            }
        }
    },
}

\definecolor{imtatlantique}{rgb}{0.005,0.715,0.867}

\title{Laplacian Networks: Bounding Indicator Function Smoothness for Neural Networks Robustness}

\author{Carlos Eduardo Rosar Kos Lassance, Vincent Gripon \\
  IMT Atlantique\\
  Brest, France \\
  \texttt{first dot lastname at imt-atlantique dot fr} \\
  \And
Antonio Ortega \\
  University of Southern California \\
  Los Angeles, USA \\
  \texttt{first dot lastname at sipi dot usc dot edu} \\
}

\begin{document}

\maketitle

\begin{abstract}
  For the past few years, Deep Neural Network (DNN) robustness has become a question of paramount importance. As a matter of fact, in sensitive settings misclassification can lead to dramatic consequences. Such misclassifications are likely to occur when facing 
  adversarial attacks, hardware failures or limitations,  
  and imperfect signal acquisition.
  To address this question, authors have proposed different approaches aiming at increasing the robustness of DNNs, such as adding regularizers or training using noisy examples.
  In this paper we propose a new regularizer built upon the Laplacian of similarity graphs obtained from the representation of training data at each layer of the DNN architecture. This regularizer penalizes large changes (across consecutive layers in the architecture) in the distance between examples of different classes, and as such enforces smooth variations of the class boundaries. 
  Since it is agnostic to the type of deformations that are expected when predicting with the DNN, the proposed regularizer can be combined with existing ad-hoc methods.
  We provide theoretical justification for this regularizer and demonstrate its effectiveness to improve robustness of DNNs on classical supervised learning vision datasets.
\end{abstract}

\section{Introduction}

Deep Neural Networks (DNNs) provide state-of-the-art performance in many challenges in machine learning~\citep{he2016identity,wu2016google}. 
Their ability to achieve good generalization is often explained by the fact they use very few priors about data~\citep{lecun2015deep}.
On the other hand, their strong dependency on data may lead to overfit to unrelevant features of the training dataset, resulting in a nonrobust classification performance.

In the literature, authors have been interested in studying the robustness of DNNs in various conditions. These conditions include:
\begin{itemize}
\item Robustness to isotropic noise, i.e., small isotropic variations of the input~\citep{mallat2016understanding}, typically meaning that the network function leads to a small Lipschitz constant.
\item Robustness to adversarial attacks, which can exploit knowledge about the network parameters or the training dataset~\citep{szegedy2013intriguing,goodfellow2014explaining}.
\item Robustness to implementation defects, which can result in only approximately correct computations~\citep{hubara2017quantized}.
\end{itemize}


To improve DNN robustness, three main families of solutions have been proposed in the literature. The first one involves enforcing smoothness, as measured by a Lipschitz constant, in the operators and having a minimum separation margin~\citep{mallat2016understanding}. A similar approach has been proposed in~\citep{cisse2017parseval}, where the authors restrict the function of the network to be contractive.
A second class of methods use intermediate representations obtained at various layers during the prediction phase~\citep{papernot2017deepknn}.
Finally, in~\citep{kurakin2016adversarial,pezeshki2016deconstructing}, the authors propose to train the network using noisy inputs so that it better generalizes to this type of noise. 
This has been shown to improve the robustness of the network to the specific type of noise used during training, but it is not guaranteed that this robustness would be extended to other types of deformations.

In this work, we introduce a new regularizer that does not focus on a specific type of deformation, but aims at increasing robustness in general. As such, the proposed regularizer can be combined with other existing methods. It is inspired by recent developments in Graph Signal Processing (GSP)~\citep{shuman2013emerging}. GSP is a mathematical framework that extends classical Fourier analysis to complex topologies described by graphs, by introducing notions of frequency for signals defined on graphs. Thus, signals that are smooth on the graph (i.e., change slowly from one node to its neighbors) will have most of their energy concentrated in the low frequencies. 

The proposed regularizer is based on constructing a series of graphs, one for each layer of the DNN architecture, where each graph captures the similarity between all training examples given their intermediate representation at that layer. Our proposed regularizer penalizes large changes in the smoothness of class indicator vectors (viewed here as graph signals) from one layer to the next. As a consequence, the distances between pairs of examples in different classes are only allowed to change slowly from one layer to the next. Note that because we use deep architectures, the regularizer does not prevent the smoothness from achieving its maximum value, but constraining the size of changes from layer to layer reduces the risk of overfitting by controlling the distance to the boundary region, as supported by experiments in Section~\ref{experiments}. 

The outline of the paper is as follows. In Section~\ref{relatedwork} we present related work. In Section~\ref{methodo} we introduce the proposed regularizer. In Section~\ref{experiments} we evaluate the performance of our proposed method in various conditions and on vision benchmarks. Section~\ref{conclusion} summarizes our conclusions.

\section{Related work}
\label{relatedwork}
DNN robustness may refer to many different problems. In this work we are mostly interested in the stability to deformations~\citep{mallat2016understanding}, or noise, which can be due to multiple factors mentioned in the introduction.
The most studied stability to deformations is in the context of adversarial attacks. It has been shown that very small imperceptible changes on the input of a trained DNN can result in missclassification of the input~\citep{szegedy2013intriguing,goodfellow2014explaining}. These works have been primordial to show that DNNs may not be as robust to deformations as the test accuracy benchmarks would have lead one to believe. Other works, such as~\citep{recht2018cifar}, have shown that DNNs may also suffer from drops in performance when facing deformations that are not originated from adversarial attacks, but simply by re-sampling the test images.


Multiple ways to improve robustness have been proposed in the literature. They range from the use of a model ensemble composed of $k$-nearest neighbors classifiers for each layer~\citep{papernot2017deepknn}, to the use of distillation as a mean to protect the network~\citep{papernot2016distillation}. Other methods introduce regularizers~\citep{gu2014towards}, control the Lipschitz constant of the network function~\citep{cisse2017parseval} or implement multiple strategies revolving around using deformations as a data augmentation procedure during the training phase~\citep{goodfellow2014explaining,kurakin2016adversarial,moosavi2016deepfool}.

Compared to these works, our proposed method can be viewed as a regularizer
that penalizes large deformations of the class boundaries throughout the network architecture, instead of focusing on a specific deformation of the input. As such, it can be combined with other mentioned strategies. Indeed, we demonstrate that the proposed method can be implemented in combination with~\citep{cisse2017parseval}, resulting in a network function such that small variations to the input lead to small variations in the decision, as in ~\citep{cisse2017parseval}, while limiting the amount of change to the class boundaries. 
Note that our approach does not require using training data affected by a specific deformation, and our results could be further improved if such data were available for training.

As for combining GSP and machine learning, this area has sparked interest recently. For example, the authors of~\citep{GriOrtGir20182} show that it is possible to detect overfitting by tracking the evolution of the smoothness of a graph containing only training set examples. Another example is in~\citep{anirudh2017influential} where the authors introduce different quantities related to GSP that can be used to extract interpretable results from DNNs. In~\citep{svoboda2018peernets} the authors exploit graph convolutional layers~\citep{bronstein2017geometric} to increase the robustness of the network.

 To the best of our knowledge, this is the first use of graph signal smoothness as a regularizer for deep neural network design.

\section{Methodology}
\label{methodo}

\subsection{Similarity preset and postset graphs}

Consider a deep neural network architecture. Such a network is obtained by assembling layers of various types. Of particular interest are layers of the form $\mathbf{x}^\ell\mapsto \mathbf{x}^{\ell+1} = h^\ell(\mathbf{W}^\ell \mathbf{x}^\ell + \mathbf{b}^\ell)$, where $h^\ell$ is a nonlinear function, typically a ReLU, $\mathbf{W}^\ell$ is the weight tensor at layer $\ell$, $\mathbf{x}^\ell$ is the intermediate representation of the input at layer $\ell$ and $\mathbf{b}^\ell$ is the corresponding bias tensor. Note that strides or pooling may be used. Assembling can be achieved in various ways: composition, concatenation, sums\dots so that we obtain a global function $f$ that associates an input tensor $\mathbf{x}^0$ to an output tensor $\mathbf{y} = f(\mathbf{x}^0)$.

When computing the output $\mathbf{y}$ associated with the input $\mathbf{x}^0$, each layer $\ell$ of the architecture processes some input $\mathbf{x}^\ell$ and computes the corresponding output $\mathbf{y}^\ell = h^\ell(\mathbf{W}^\ell \mathbf{x}^\ell + \mathbf{b}^\ell)$. 
For a given layer $\ell$ and a batch of $b$ inputs $\mathcal{X}=\{\mathbf{x}_1,\dots,\mathbf{x}_b\}$, we can obtain two sets $\mathcal{X}^\ell = \{\mathbf{x}^\ell_1,\dots,\mathbf{x}^\ell_b\}$, called the \emph{preset}, and $\mathcal{Y}^\ell = \{\mathbf{y}^\ell_1,\dots,\mathbf{y}^\ell_b\}$, called the \emph{postset}.

Given a similarity measure $s$ on tensors, from a preset we can build the similarity preset matrix: $\mathbf{M}^\ell_{pre}[i,j] = s(\mathbf{x}_i^\ell, \mathbf{x}_j^\ell), \forall 1\leq i,j \leq b$, where $\mathbf{M}[i,j]$ denotes the element at line $i$ and column $j$ in $\mathbf{M}$. The postset matrix is defined similarly.

Consider a similarity (either preset or postset) matrix $\mathbf{M}^\ell$. This matrix can be used to build a $k$-nearest neighbor similarity weighted graph $G^\ell = \langle V, \mathbf{A}^\ell\rangle$, where $V = \{1,\dots,b\}$ is the set of vertices and $\mathbf{A}^\ell$ is the weighted adjacency matrix defined as:
\begin{equation}\displaystyle{\mathbf{A}^\ell[i,j] = \left\{\begin{array}{ll}\mathbf{M}^\ell[i,j] & $if $ \mathbf{M}^\ell[i,j] \in \arg\max_{i'\neq j}{(\mathbf{M}^\ell[i',j],k)}\\&\;\;\;\;\;\;\;\;\;\;\;\;\;\;\;\;\bigcup \arg\max_{j' \neq i}{(\mathbf{M}^\ell[i,j'],k)}\\0 &$otherwise$\end{array}\right.},\forall i,j \in V,\end{equation}where $\arg\max_{i}(a_i, k)$ denotes the indices of the $k$ largest elements in $\{a_1,\dots,a_b\}$.
Note that by construction $\mathbf{A}^\ell$ is symmetric.

\subsection{Smoothness of label signals}

Given a weighted graph $G^\ell = \langle V, \mathbf{A}^\ell\rangle$, we call Laplacian of $G^\ell$ the matrix $\mathbf{L}^\ell = \mathbf{D}^\ell - \mathbf{A}^\ell$, where $\mathbf{D}^\ell$ is the diagonal matrix such that: $\mathbf{D}^\ell[i,i] = \sum_{j}{\mathbf{A}^\ell[i,j]}, \forall i \in V$.
Because $\mathbf{L}^\ell$ is symmetric and real-valued, it can be written:
\begin{equation} \mathbf{L}^\ell = \mathbf{F}^\ell \mathbf{\Lambda}^\ell \mathbf{F}^{\ell\top},\end{equation}
where $\mathbf{F}$ is orthonormal and contains eigenvectors of $\mathbf{L}^\ell$ as columns, $\mathbf{F}^\top$ denotes the transpose of $\mathbf{F}$, and $\mathbf{\Lambda}$ is diagonal and contains eigenvalues of $\mathbf{L}^\ell$ is ascending order. Note that the constant vector $\mathbf{1}\in\mathbb{R}^b$ is an eigenvector of $\mathbf{L}^\ell$ corresponding to eigenvalue $0$. Moreover, all eigenvalues of $\mathbf{L}^\ell$ are nonnegative. Consequently, $\mathbf{1}/\sqrt{n}$ can be chosen as the first column in $\mathbf{F}$.

Consider a vector $\mathbf{s}\in\mathbb{R}^b$, we define $\hat{\mathbf{s}}$ the Graph Fourier Transform (GFT) of $\mathbf{s}$ on $G^\ell$ as \citep{shuman2013emerging}:  \begin{equation}\hat{\mathbf{s}} = \mathbf{F}^\top \mathbf{s}.\end{equation}
Because the order of the eigenvectors is chosen so that the corresponding eigenvalues are in ascending order, if only the first few entries of $\hat{\mathbf{s}}$ are nonzero that indicates that $\mathbf{s}$ is low frequency (smooth). In the extreme case where only the first entry of $\hat{\mathbf{s}}$ is nonzero we have that $\mathbf{s}$ is constant (maximum smoothness).
More generally, smoothness $\sigma^\ell(\mathbf{s})$ of a signal $\mathbf{s}$ can be measured using the quadratic form of the Laplacian:
\begin{equation}\sigma^\ell(\mathbf{s}) = \mathbf{s}^\top \mathbf{L}^\ell \mathbf{s} = \sum_{i,j = 1}^{b}{\mathbf{A}^\ell[i,j](\mathbf{s}[i] - \mathbf{s}[j])^2} = \sum_{i = 1}^{b}{\mathbf{\Lambda}^\ell[i,i] \hat{\mathbf{s}}[i]^2},\label{smoothness}\end{equation}
where we note that $\mathbf{s}$ is smoother when $\sigma^\ell(\mathbf{s})$ is smaller. 

In this paper we are particularly interested in smoothness of the label signals. We call \emph{label signal} $\mathbf{s}_c$ associated with class $c$ a binary ($\{0,1\}$) vector whose nonzero coordinates are the ones corresponding to input vectors of class $c$. In other words, $\mathbf{s}_c[i] = 1 \Leftrightarrow (\mathbf{x}_i$ is in class $c ), \forall 1\leq i\leq b$.

Denote $u$ the last layer of the architecture: $ \mathbf{y}^u_i = \mathbf{y}_i, \forall i$. Note that in typical settings, where outputs of the networks are one-hot-bit encoded and no regularizer is used, at the end of the learning process it is expected that $\mathbf{y}_i^\top \mathbf{y}_j \approx 1$ if $i$ and $j$ belong to the same class, and $\mathbf{y}_i^\top \mathbf{y}_j \approx 0$ otherwise.

Thus, assuming that cosine similarity is used to build the graph, the last layer smoothness for all $c$ would be $ \sigma^u_{post}(\mathbf{s}_c) \approx 0$, since edge weights between nodes having different labels will be close to zero given  Equation~(\ref{smoothness}). More generally, smoothness of $\mathbf{s}_c$ at the preset or postset of a given layer measures the average similarity between examples in class $c$ and examples in other classes ($\sigma(\mathbf{s}_c)$ decreases as the weights of edges connecting nodes in different classes decrease). Because the last layer can achieve $\sigma(\mathbf{s}_c)\approx 0$, we expect the  smoothness metric $\sigma$ at each layer to decrease as we go deeper in the network. 
Next we introduce a regularization strategy that limits how much $\sigma$ can decrease from one layer to the next and can even prevent the last layer from achieving $\sigma(\mathbf{s}_c)=0$. This will be shown to improve generalization and robustness.  
The theoretical motivation for this choice is discussed in Section~\ref{motivation}.

\subsection{Proposed regularizer}

\subsubsection{Definition}

We propose to measure the deformation induced by a given layer $\ell$ in the relative positions of examples by computing the difference between label signal smoothness before and after the layer, averaged over all labels:
\begin{equation}\delta_\sigma^\ell = \left|\sum_c{\left[\sigma^\ell_{post}(\mathbf{s}_c) - \sigma^\ell_{pre}(\mathbf{s}_c)\right]}\right|.\end{equation}
These quantities are used to regularize modifications made to each of the layers during the learning process.

\emph{Remark 1:} 
Since we only consider label signals, we solely depend on the similarities between examples that belong to distinct classes. In other words, the regularizer only focuses on the boundary region, and does not vary if the distance between examples of the same label grows or shrinks. This is because forcing similarities between examples of a same class to evolve slowly could prevent the network to train appropriately. 

\emph{Remark 2:}
Compared with~\citep{cisse2017parseval}, there are three key differences that characterize the proposed regularizer:
\begin{enumerate}
    \item Not all pairwise distances are taken into account in the regularization; only distances between examples corresponding to different classes play a role in the regularization.
    \item We allow a limited amount of both contraction and dilation of the metric space. Experimental work (e.g.~\citep{GriOrtGir20182,papernot2017deepknn}) has shown that the evolution of metric spaces across DNN layers is complex, and thus restricting ourselves to contractions only could lead to lower overall performance.
    \item The proposed criterion is an average (sum) over all distances, rather than a stricter criterion (e.g. Lipschitz), which would force each pair of vectors $(\mathbf{x}_i, \mathbf{x}_j)$ to obey the constraint.
\end{enumerate}

\textbf{Illustrative example:}

\begin{figure}
    \centering
    \begin{tikzpicture}[thick]
    \node at (-3,0) {\textbf{Initial problem:}};
    \begin{scope}[scale=1.5,xshift=-1cm]
    \draw[fill=white]
    (0.2,0) circle (0.07)
    (0.4,0) circle (0.07)
    (0.6,0) circle (0.07)
    (3.1,0) circle (0.07)
    (3.3,0) circle (0.07)
    (3.6,0) circle (0.07);
    \draw
    (1.5,0.07) -- (1.64,-0.07)
    (1.64,0.07) -- (1.5,-0.07)
    (1.8,0.07) -- (1.94,-0.07)
    (1.94,0.07) -- (1.8,-0.07)
    (2.1,0.07) -- (2.24,-0.07)
    (2.24,0.07) -- (2.1,-0.07);
    \draw[red, >=stealth', |-|]
    (0.9, 0) -- (1.3, 0);
    \draw[red, >=stealth', |-|]
    (2.44, 0) -- (2.84, 0)
    ;
    \draw[>=stealth',->, black!50!white]
    (0,-0.3) -- (4,-0.3);
    \foreach \i in {0,...,11}{
        \draw[black!50!white]({0.33*\i},-0.25) -- ({0.33*\i},-0.35);
    }
    \node(abr) at (1.8,1) {Class domains boundary};
    \path[=>stealth',->]
    (abr) edge[bend left] (2.64, 0.2)
    (abr) edge[bend right] (1.0, 0.2);
    \end{scope}
    \node[text width=5cm] at (-3,-2) {\textbf{i) No regularization:}};
    \begin{scope}[scale=1.5,xshift=-4cm, yshift=-2cm]
    \draw[>=stealth',->,black!50!white] plot [smooth, tension=2]
    coordinates {(4,-0.5) (0,-0.125) (4,0.25)}
    [arrow inside={end=|,opt={scale=.5}}{0.03,0.06,0.09,0.25, 0.45,0.48,0.52,0.55, 0.75, 0.91, 0.94,0.97}];
    
    \draw[fill=white]
    (3.3,-0.3) circle (0.07)
    (3.55,-0.3) circle (0.07)
    (3.8,-0.3) circle (0.07)
    (3.3,0.4) circle (0.07)
    (3.55,0.4) circle (0.07)
    (3.8,0.4) circle (0.07);
    
    \begin{scope}[xshift=-0.7cm, yshift=-0.10cm]
    \draw
    (0.5,0.07) -- (0.64,-0.07)
    (0.64,0.07) -- (0.5,-0.07);
    \end{scope} 
    \begin{scope}[xshift=-0.5cm, yshift=-0.45cm]
    \draw
    (0.5,0.07) -- (0.64,-0.07)
    (0.64,0.07) -- (0.5,-0.07);
    \end{scope} 
    \begin{scope}[xshift=-0.5cm, yshift=0.20cm]
    \draw
    (0.5,0.07) -- (0.64,-0.07)
    (0.64,0.07) -- (0.5,-0.07);
    \end{scope} 
    \draw[red, |-|]
    (1.15, -0.3) -- (2.65, -0.35);
    \draw[red, |-|]
    (1.15, 0.28) -- (2.65, 0.35);
    \end{scope}
    \node[text width=5cm] at (4.5,-2) {\textbf{ii) Proposed regularization:}};
    \begin{scope}[scale=1.5, xshift=1cm, yshift=-2cm]
    \draw[>=stealth',->,black!50!white] plot [smooth, tension=2]
    coordinates {(4,-0.5) (0,-0.125) (4,0.25)}
    [arrow inside={end=|,opt={scale=.5}}{0.08,	0.16,	0.24,	0.32,	0.40,	0.48,	0.56,	0.64,	0.72,	0.80,	0.88,	0.96}];
    
    \draw[fill=white]
    (2.0,-0.3) circle (0.07)
    (2.7,-0.3) circle (0.07)
    (3.3,-0.3) circle (0.07)
    (2.4,0.4) circle (0.07)
    (3.0,0.4) circle (0.07)
    (3.6,0.4) circle (0.07);
    
    \begin{scope}[xshift=-0.7cm, yshift=-0.10cm]
    \draw
    (0.5,0.07) -- (0.64,-0.07)
    (0.64,0.07) -- (0.5,-0.07);
    \end{scope} 
    \begin{scope}[xshift=0.1cm, yshift=-0.45cm]
    \draw
    (0.5,0.07) -- (0.64,-0.07)
    (0.64,0.07) -- (0.5,-0.07);
    \end{scope} 
    \begin{scope}[xshift=0.13cm, yshift=0.25cm]
    \draw
    (0.5,0.07) -- (0.64,-0.07)
    (0.64,0.07) -- (0.5,-0.07);
    \end{scope} 
    \draw[red, |-|]
    (1.1, -0.28) -- (1.7, -0.32);
    \draw[red, |-|]
    (1.4, 0.29) -- (2.0, 0.33)
    ;
    \end{scope}

    \end{tikzpicture}
    \caption{Illustration of the effect of our proposed regularizer. In this example, the goal  is to classify circles and crosses (top).  Without use of regularizers (bottom left), the resulting embedding may considerably stretch the boundary regions (as illustrated by the irregular spacing between the tics). Forcing small variations of smoothness of label signals (bottom right), we ensure the topology is not dramatically changed  in the boundary regions.}
    \label{illustrativeexample}
\end{figure}
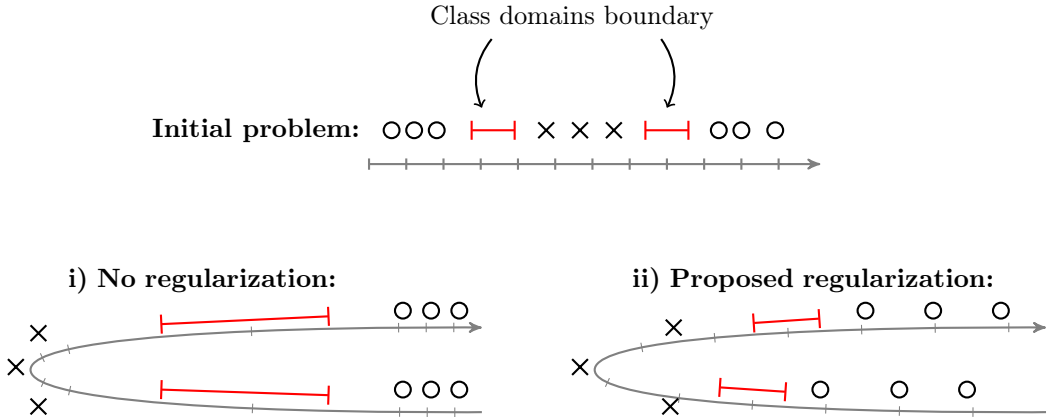

In Figure~\ref{illustrativeexample} we depict a toy illustrative example to motivate   the proposed regularizer. We consider here a one-dimensional two-class problem. To linearly separate circles and crosses, it is necessary to group all circles. Without regularization (setting i)), the resulting embedding is likely to increase considerably  the distance between examples and 
the size of the boundary region between classes. In contrast, by penalizing large variations of the smoothness of label signals (setting ii)), the average distance between circles and crosses must be preserved in the embedding domain, resulting in a more precise control of distances within the boundary region. 

\subsection{Motivation: label signal bandwidth and powers of the Laplacian}
\label{motivation}

Recent work \citep{anis2017sampling} develops an asymptotic analysis of the bandwidth of label signals, $BW(\mathbf{s})$, where bandwidth is defined as the highest non-zero graph frequency of $\mathbf{s}$, i.e., the nonzero entry of $\hat{\mathbf{s}}$ with the highest index. An estimate of the bandwidth can be obtained by computing:
\begin{equation}
BW_m(\mathbf{s}) = \left( \frac{\mathbf{s}^{\top}\mathbf{L}^m \mathbf{s}}{\mathbf{s}^{\top}\mathbf{s}} \right)^{(1/m)}
\end{equation}
for large $m$. 
This can be viewed as a generalization of the smoothness metric of (\ref{smoothness}). \citep{anis2017sampling} shows that, as the number of labeled points $\mathbf{x}$ (assumed drawn from a distribution $p(\mathbf{x})$)  grows asymptotically,  the bandwidth of the label signal converges in probability to the supremum of $p(\mathbf{x})$ in the region of overlap between classes. This  motivates our work in three ways. 

First, it provides theoretical justification to use $\sigma^\ell(\mathbf{s})$ for regularization, since lower values of $\sigma^\ell(\mathbf{s})$ are indicative of better separation between classes. Second, the asymptotic analysis suggests that using higher powers of the Laplacian would lead to better regularization, since estimating bandwidth using $BW_m(\mathbf{s})$ becomes increasingly accurate as $m$ increases. Finally, this regularization can be seen to be protective against overfitting by preventing $\sigma^\ell(\mathbf{s})$ from decreasing ``too fast''. For most problems of interest, given a sufficiently large amount of labeled data available, it would be reasonable to expect the bandwidth of $\mathbf{s}$ not to be arbitrarily small, because the classes cannot be exactly separated, and thus a network that reduces the bandwidth too much can result in overfitting. 

\subsection{Analysis of the Laplacian powers}

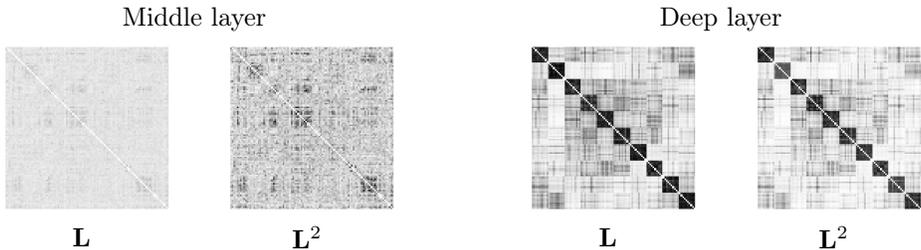
\begin{figure}[t]
    \centering
    \begin{tikzpicture}
    \node at (1.5, 1.5) {Middle layer};
    \node at (0,0){\includegraphics[width=2.5cm]{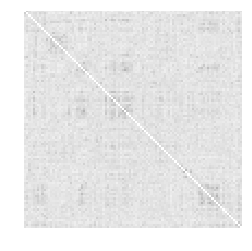}};
    \node at (0,-1.4) {$\mathbf{L}$};
    \node at (3,-1.4) {$\mathbf{L}^2$};
    \node at (7,-1.4) {$\mathbf{L}$};
    \node at (10,-1.4) {$\mathbf{L}^2$};
    \node at (7,0) {\includegraphics[width=2.5cm]{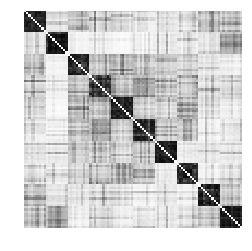}};
    \node at (3,0) {\includegraphics[width=2.5cm]{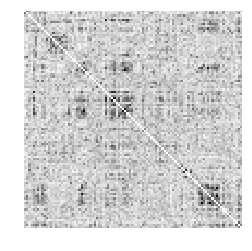}};
    \node at (8.5, 1.5) {Deep layer};
    \node at (10,0) {\includegraphics[width=2.5cm]{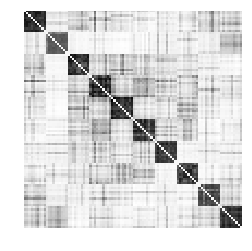}};
    
    \end{tikzpicture}
     \vspace{-0.35cm}
    \caption{Sample of a Laplacian and squared Laplacian of similarity graphs in a trained vanilla architecture. Examples of the batch have been ordered so that those belonging to a same class are consecutive. Dark values correspond to high similarity.}
   \label{laplacianpowers}
\end{figure}

In Figure~\ref{laplacianpowers} we depict the Laplacian and squared Laplacian of similarity graphs obtained at different layers in a trained vanilla architecture. On the deep layers, we can clearly see blocks corresponding to the classes, while the situation in the middle layer is not as clear. This figure illustrates how using the squared Laplacian helps modifying the distances to improve separation. 
Note that we normalize the squared Laplacian values by dividing them by the highest absolute value. 

\begin{figure}[t]


 \begin{center}
   \begin{tikzpicture}
     \begin{scope}[yscale=.5,xscale=.5]
       \begin{axis}[
           xlabel=Layer depth,
           ylabel=Smoothness,legend pos=north west,
           title=$\mathbf{L}^2$]
         \addplot table {values/vanilla_squared.txt};
         \addlegendentry{Vanilla}
         \addplot table {values/reg_squared.txt};
         \addlegendentry{Regularization with $\mathbf{L}$}
         \addplot table {values/squared_squared.txt};
         \addlegendentry{Regularization with $\mathbf{L}^2$}
       \end{axis}
     \end{scope}
      \begin{scope}[xshift=-6.5cm, xscale=.5, yscale=.5]
       \begin{axis}[
           xlabel=Layer depth,
           ylabel=Smoothness,legend pos=north east,
           title=$\mathbf{L}$]
         \addplot table {values/vanilla.txt};
         \addlegendentry{Vanilla}
         \addplot table {values/reg.txt};
         \addlegendentry{Regularization with $\mathbf{L}$}
         \addplot table {values/squared.txt};
         \addlegendentry{Regularization with $\mathbf{L}^2$}
       \end{axis}
     \end{scope}
   \end{tikzpicture}
 \end{center}
    \vspace{-0.35cm}
    \caption{Evolution of smoothness of label signals as a function of layer depth, and for various regularizers and choice of $m$, the power of the Laplacian matrix.}
    \label{smoothness_evolution}
\end{figure}
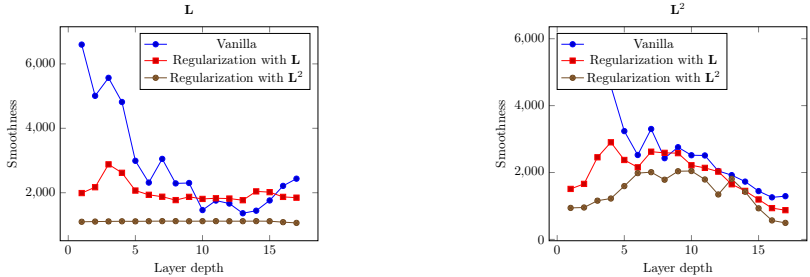

In Figure~\ref{smoothness_evolution}, we plot the average evolution of smoothness of label signals over 100 batches, as a function of layer depth in the architecture, and for different choices of the regularizer. In the left part, we look at smoothness measures using the Laplacian. In the right part, we use the squared Laplacian. We can clearly see the effectiveness of the regularizer in enforcing small variations of smoothness across the architecture. Note that for model regularized with $\mathbf{L}^2$, changes in smoothness measured by $\mathbf{L}$ are not easy to see. This seems to suggest that some of the gains achieved via $\mathbf{L}^2$ regularization come in making changes that would be ``invisible'' when looking at the layers from the perspective of $\mathbf{L}$ smoothness. The same normalization from Figure~\ref{laplacianpowers} is used for $\mathbf{L}^2$.

\section{Experiments}
\label{experiments}

In the following paragraphs we evaluate the proposed method using various tests. We use the well known CIFAR-10~\citep{krizhevsky2009learning} dataset made of tiny images. As far as the DNN is concerned, we use the same PreActResNet~\citep{he2016identity} architecture for all tests, with 18 layers. All inputs, including those on the test set, are normalized based on the mean and standard deviation of the images of the \emph{training} set. Discussion about the implementation of the Parseval training, hyperparameters and more details can be found at the Appendix.



We depict the obtained results using box plots where data is aggregated from 10 different networks corresponding to different random seeds and batch orders. In the first experiment (left most plot) in Figure~\ref{gaussian}, we plot the baseline accuracy of the models on the clean test set (no deformation is added at this point). These experiments agree with the claim from~\citep{cisse2017parseval} where the authors show that they are able to increase the performance of the network on the clean test set. We observe that our proposed method leads to a minor decrease of performance on this test. However, we see in the following experiments that this is mitigated with increased robustness to deformations.


\subsection{Isotropic deformation}

\begin{figure}[t]
    \begin{tikzpicture}

    \node at (-2.2cm,1cm) {$SNR \approx \infty$};
    \node at (2.cm,1cm) {$SNR \approx 20$};
    \node at (6.cm,1cm) {$SNR \approx 15$};
    \begin{scope}[xscale=0.55,yscale=0.4,xshift=-7.5cm,yshift=-4.5cm]
      \begin{axis}
        [xmin=0,xmax=100,
        ytick={4,3,2,1},
        yticklabels={Vanilla, Parseval, Proposed Method, Proposed + Parseval},
        xlabel={Test set accuracy}
        ]
        \addplot+[
        boxplot prepared={
          median=87.35,
          upper quartile=87.47,
          lower quartile=87.00,
          upper whisker=88.18,
          lower whisker=86.30
        },
        ] coordinates {};
        \addplot+[
        boxplot prepared={
          median=86.78,
          upper quartile=86.86,
          lower quartile=86.49,
          upper whisker=87.48,
          lower whisker=85.94
        },
        ] coordinates {};
        \addplot+[
        boxplot prepared={
          lower quartile=89.62,
          median=89.75,
          upper quartile=89.84,
          upper whisker=90.17,
          lower whisker=89.29
        },
        ] coordinates {};
        \addplot+[
        boxplot prepared={
          median=88.09,
          upper quartile=88.31,
          lower quartile=87.92,
          upper whisker=88.90,
          lower whisker=87.335
        },
        ] coordinates {};
      \end{axis}
    \end{scope}
    \begin{scope}[xscale=0.55,yscale=0.4,xshift=0cm,yshift=-4.5cm]
      \begin{axis}
        [xmin=0,xmax=100,ytick={4,3,2,1},yticklabels={, , , },
        xlabel={Test set accuracy}]
        \addplot+[
        boxplot prepared={
          lower quartile=78.29,
          median=78.96,
          upper quartile=79.53,
          upper whisker=81.39,
          lower whisker=76.43
        },] coordinates {};
        \addplot+[
        boxplot prepared={
          lower quartile=76.27,
          median=77.0,
          upper quartile=77.88,
          upper whisker=80.30,
          lower whisker=73.86
        },] coordinates {};
        \addplot+[
        boxplot prepared={
          lower quartile=65.84,
          median=67.50,
          upper quartile=69.02,
          upper whisker=73.79,
          lower whisker=61.07
        },
        ] coordinates {};
        \addplot+[
        boxplot prepared={
          lower quartile=66.37,
          median=67.81,
          upper quartile=69.05,
          upper whisker=73.07,
          lower whisker=62.35
        },
        ] coordinates {};
      \end{axis}
    \end{scope}

    \begin{scope}[xscale=0.55,yscale=0.4,xshift=7.5cm,yshift=-4.5cm]
      \begin{axis}
        [xmin=0,xmax=100,ytick={4,3,2,1},yticklabels={, , , },
        xlabel={Test set accuracy}]
        \addplot+[
        boxplot prepared={
          lower quartile=48.69,
          median=50.94,
          upper quartile=53.24,
          upper whisker=60.07,
          lower whisker=41.87
        },] coordinates {};
        \addplot+[
        boxplot prepared={
          lower quartile=41.02,
          median=44.26,
          upper quartile=47.10,
          upper whisker=56.22,
          lower whisker=31.9
        },] coordinates {};
        \addplot+[
        boxplot prepared={
          lower quartile=29.41,
          median=32.3,
          upper quartile=35.43,
          upper whisker=44.46,
          lower whisker=20.38
        },] coordinates {};
        \addplot+[
        boxplot prepared={
          lower quartile=32.28,
          median=35.50,
          upper quartile=38.64,
          upper whisker=48.18,
          lower whisker=22.74
        },] coordinates {};
      \end{axis}
    \end{scope}

    \end{tikzpicture}
    \vspace{-0.35cm}
    \caption{Test set accuracy under Gaussian Noise with varying signal-to-noise ratio.}
    \label{gaussian}
\end{figure}
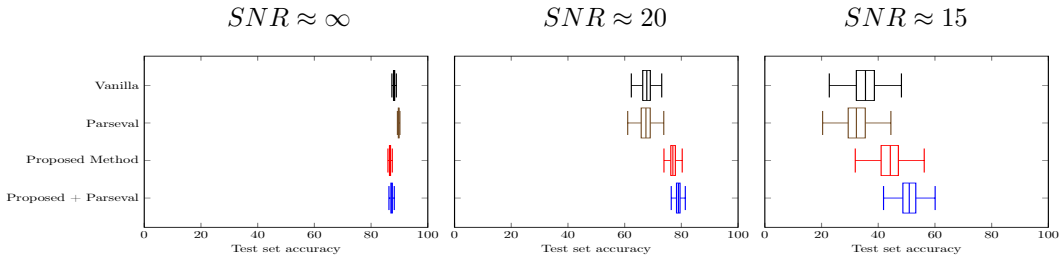

In this scenario we evaluate the robustness of the network function to small isotropic variations of the input. We generate 40 different deformations using random variables $\mathcal{N}(0,0.25)$ which are added to the test set inputs. Note that they are scaled so that $SNR \approx 15$ and $SNR \approx 20$. The middle and right-most plots from Figure~\ref{gaussian} show that the proposed method increases the robustness of the network to isotropic deformations. Note that in both scenarios the best results are achieved by combining Parseval training and our proposed method (lower-most box on both figures).

\subsection{Adversarial Robustness}

We next evaluate robustness to adversarial inputs, which are specifically built to fool the network function. Such adversarial inputs can be generated and evaluated in multiple ways. Here we implement two approaches: first a mean case of adversarial noise, where the adversary can only use one forward and one backward pass to generate the deformations, and second a worst case scenario, where the adversary can use multiple forward and backward passes to try to find the smallest deformation that will fool the network.

\begin{figure}[t]
    \begin{tikzpicture}
    \node at (-2.2cm,1cm) {FGSM after norm};
    \node at (2cm,1cm) {FGSM before norm};
    \node at (6cm,1cm) {DeepFool};
    \begin{scope}[xscale=0.55,yscale=0.4,xshift=-7.5cm,yshift=-4.5cm]
      \begin{axis}
        [ytick={4,3,2,1},yticklabels={Vanilla, Parseval, Proposed Method, Proposed + Parseval},xmin=0,xmax=100,
        xlabel={Test set accuracy}]
        \addplot+[
        boxplot prepared={
          lower quartile=50.44,
          median=50.71,
          upper quartile=51.33,
          upper whisker=52.665,
          lower whisker=49.105
        },] coordinates {};
        \addplot+[
        boxplot prepared={
          lower quartile=49.74,
          median=50.51,
          upper quartile=51.00,
          upper whisker=52.89,
          lower whisker=47.85
        },] coordinates {};
        \addplot+[
        boxplot prepared={
          lower quartile=44.47,
          median=45.05,
          upper quartile=45.68,
          upper whisker=47.495,
          lower whisker=42.655
        },] coordinates {};
        \addplot+[
        boxplot prepared={
          lower quartile=33.25,
          median=33.69,
          upper quartile=34.30,
          upper whisker=35.875,
          lower whisker=31.675
        },] coordinates {};
      \end{axis}
    \end{scope}

    \begin{scope}[xscale=0.55,yscale=0.4,xshift=0cm,yshift=-4.5cm]
      \begin{axis}[ytick={4,3,2,1},yticklabels={, , , },xmin=0,xmax=100,
        xlabel={Test set accuracy}]
        \addplot+[
        boxplot prepared={
          lower quartile=27.83,
          median=28.50,
          upper quartile=28.99,
          upper whisker=30.73,
          lower whisker=26.09
        },] coordinates {};
        \addplot+[
        boxplot prepared={
          lower quartile=24.33,
          median=25.44,
          upper quartile=25.88,
          upper whisker=28.205,
          lower whisker=22.005
        },] coordinates {};
        \addplot+[
        boxplot prepared={
          lower quartile=25.98,
          median=26.35,
          upper quartile=26.56,
          upper whisker=27.43,
          lower whisker=25.11
        },] coordinates {};
        \addplot+[
        boxplot prepared={
          lower quartile=16.53,
          median=17.01,
          upper quartile=17.95,
          upper whisker=20.08,
          lower whisker=14.4
        },] coordinates {};
      \end{axis}
    \end{scope}

    \begin{scope}[xscale=0.55,yscale=0.4,xshift=7.5cm,yshift=-4.5cm]
      \begin{axis}
        [ytick={4,3,2,1},yticklabels={, , , },xmin=0,
        xlabel={Mean $\mathcal{L}_2$ pixel distance}]
        \addplot+[
        boxplot prepared={
          lower quartile=0.000069,
          median=0.000071,
          upper quartile=0.000071,
          upper whisker=7.4E-05,
          lower whisker=6.6E-05
        },] coordinates {};
        \addplot+[
        boxplot prepared={
          lower quartile=0.000067,
          median=0.000069,
          upper quartile=0.000070,
          upper whisker=7.45E-05,
          lower whisker=6.25E-05
        },] coordinates {};
        \addplot+[
        boxplot prepared={
          lower quartile=0.000058,
          median=0.000060,
          upper quartile=0.000061,
          upper whisker=6.55E-05,
          lower whisker=5.35E-05
        },] coordinates {};
        \addplot+[
        boxplot prepared={
          lower quartile=0.000044,
          median=0.000045,
          upper quartile=0.000045,
          upper whisker=4.65E-05,
          lower whisker=4.25E-05
        },] coordinates {};
      \end{axis}
    \end{scope}

    \end{tikzpicture}
    \vspace{-0.35cm}
    \caption{Robustness against an adversary. Measured by the test set accuracy under FGSM attack in the left and center plots and by the mean $\mathcal{L}_2$ pixel distance needed to fool the network using DeepFool on the right plot.}
    \label{adversarialNoise1}
\end{figure}
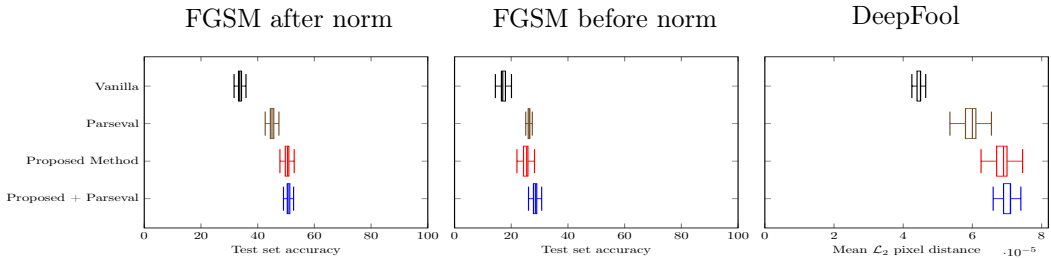

For the first approach, we add the scaled gradient sign (FGSM attack) on the input~\citep{kurakin2016adversarial}, so that we obtain a target $SNR$. Results are depicted in the left and center plots of Figure~\ref{adversarialNoise1}. In the left plot the noise is added after normalizing the input whereas on the middle plot it is added before normalizing. As in the isotropic noise case, a combination of the Parseval method and our proposed approach achieves maximum robustness.

In regards to the second approach, where a worst case scenario is considered, we use the Foolbox toolbox~\citep{rauber2017foolbox} implementation of DeepFool ~\citep{moosavi2016deepfool}. The conclusions are similar (right plot of Figure~\ref{adversarialNoise1})  to those obtained for the first adversarial attack approach.

\subsection{Implementation robustness}

Finally, in a third series of experiments we evaluate the robustness of the network functions to faulty implementations. As a result, approximate computations are made during the test phase that consist of random erasures of the memory (dropout) or quantization of the weights~\citep{hubara2017quantized}.

\begin{figure}[t]
    \begin{tikzpicture}

    \node at (-2.2cm,0.75cm) {$25\%$ dropout};
    \node at (2cm,0.75cm) {$40\%$ dropout};
    \node at (6.cm,0.75cm) {$5$ bit quantization};
    \begin{scope}[xscale=0.55,yscale=0.4,xshift=-7.5cm,yshift=-4.5cm]
      \begin{axis}
        [ytick={4,3,2,1},yticklabels={Vanilla, Parseval, Proposed Method, Proposed + Parseval},xmin=0,xmax=100,
        xlabel={Test set accuracy}]
        \addplot+[
        boxplot prepared={
          lower quartile=46.88,
          median=47.15,
          upper quartile=47.88,
          upper whisker=49.38,
          lower whisker=45.38
        },] coordinates {};
        \addplot+[
        boxplot prepared={
          lower quartile=65.65,
          median=66.57,
          upper quartile=67.57,
          upper whisker=70.45,
          lower whisker=62.77
        },] coordinates {};
        \addplot+[
        boxplot prepared={
          lower quartile=47.86,
          median=52.99,
          upper quartile=53.63,
          upper whisker=62.285,
          lower whisker=39.205
        },] coordinates {};
        \addplot+[
        boxplot prepared={
          lower quartile=44.22,
          median=44.99,
          upper quartile=48.02,
          upper whisker=53.72,
          lower whisker=38.52
        },] coordinates {};
      \end{axis}
    \end{scope}

    \begin{scope}[xscale=0.55,yscale=0.4,xshift=0cm,yshift=-4.5cm]
      \begin{axis}
        [ytick={4,3,2,1},yticklabels={, , , },xmin=0,xmax=100,
        xlabel={Test set accuracy}]
        \addplot+[
        boxplot prepared={
          lower quartile=14.19,
          median=14.87,
          upper quartile=15.69,
          upper whisker=17.94,
          lower whisker=11.94
        },] coordinates {};
        \addplot+[
        boxplot prepared={
          lower quartile=33.01,
          median=33.93,
          upper quartile=34.99,
          upper whisker=37.96,
          lower whisker=30.04
        },] coordinates {};
        \addplot+[
        boxplot prepared={
          lower quartile=16.03,
          median=16.87,
          upper quartile=17.30,
          upper whisker=19.205,
          lower whisker=14.125
        },] coordinates {};
        \addplot+[
        boxplot prepared={
          lower quartile=22.56,
          median=23.67,
          upper quartile=24.81,
          upper whisker=28.185,
          lower whisker=19.185
        },] coordinates {};
      \end{axis}
    \end{scope}

    \begin{scope}[xscale=0.55,yscale=0.4,xshift=7.5cm,yshift=-4.5cm]
      \begin{axis}
        [ytick={4,3,2,1},yticklabels={, , , },xmin=0,xmax=100,
        xlabel={Test set accuracy}]
        \addplot+[
        boxplot prepared={
          lower quartile=20.37,
          median=26.08,
          upper quartile=30.66,
          upper whisker=46.095,
          lower whisker=4.935
        },] coordinates {};
        \addplot+[
        boxplot prepared={
          lower quartile=48.13,
          median=52.29,
          upper quartile=55.35,
          upper whisker=66.18,
          lower whisker=37.3
        },] coordinates {};
        \addplot+[
        boxplot prepared={
          lower quartile=36.06,
          median=44.63,
          upper quartile=48.27,
          upper whisker=66.585,
          lower whisker=17.745
        },] coordinates {};
        \addplot+[
        boxplot prepared={
          lower quartile=14.31,
          median=19.20,
          upper quartile=21.18,
          upper whisker=31.485,
          lower whisker=4.005
        },] coordinates {};
      \end{axis}
    \end{scope}

    \end{tikzpicture}

    \caption{Test set accuracy under different types of implementation related noise.}
    \label{dropout_quantized}
\end{figure}
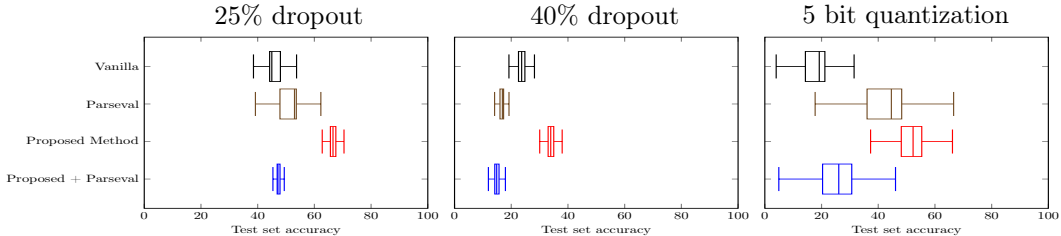

In the dropout case, we compute the test set accuracy when the network has a probability of either $25\%$ or $40\%$ of dropping a neuron's value after each block. We run each experiment $40$ times. The results are depicted in the left  and center plots of Figure~\ref{dropout_quantized}. It is interesting to note that the Parseval trained functions seem to drop in performance as soon as we reach $40\%$ probability of dropout, providing an average accuracy smaller than the vanilla networks. In contrast, the proposed method is the most robust to these perturbations.

For the quantization of the weights, we consider a scenario where the network size in memory has to be shrink 6 times. We therefore quantize the weights of the networks to 5 bits (instead of 32) and re-evaluate the test set accuracy. The right plot of Figure~\ref{dropout_quantized} shows that the proposed method is providing a better robustness to this kind of deformation than the tested counterparts.

\section{Conclusion}
\label{conclusion}

In this paper we have introduced a new regularizer that enforces small variations of the smoothness of label signals on similarity graphs obtained at intermediate layers of a deep neural network architecture.
We have empirically shown with our tests that it can lead to improved robustness in various conditions compared to existing counterparts. We also demonstrated that combining the proposed regularizer with existing methods can result in even better robustness for some conditions.

Future work includes a more systematic study of the effectiveness of the method with regards to other datasets, models and deformations. We believe that for the first two points it should not be a problem given~\citep{moosavi2017universal,papernot2016transferability} where the authors argue that adversarial noise is transferable between models and datasets.

One possible extension of the proposed method is to use it in a fine-tuning stage, combined with different techniques already established on the literature. An extension using a combination of input barycenter and class barycenter signals instead of the class signal could be interesting as that would be comparable to~\citep{zhang2017mixup}. In the same vein, using random signals could be beneficial for semi-supervised or unsupervised learning challenges.








\bibliography{iclr_2019}
\bibliographystyle{unsrtnat}

\appendix

\section{Parseval Training and implementation}

We compare our results with those obtained using the method described in~\citep{cisse2017parseval}. There are three modifications to the normal training procedure: orthogonality constraint, convolutional renormalization and convexity constraint.

For the orthogonality constraint we enforce \emph{Parseval tightness} \citep{kovavcevic2008introduction} as a layer-wise regularizer: \begin{equation}R_\beta(W^\ell) = \frac{\beta}{2}\|W^{\ell\top} W^\ell - I\|^2_2,\end{equation}
where $W_\ell$ is the weight tensor at layer $\ell$. This function can be approximately optimized with gradient descent by doing the operation: \begin{equation}W^\ell \leftarrow (1 + \beta)W^\ell - \beta W^\ell W^{\ell\top} W^\ell.\end{equation}. Given that our network is smaller we can apply the optimization to the entirety of the $W$, instead of $30\%$ as per the original paper, this increases the strength of the Parseval tightness.

For the convolutional renormalization, each matrix $W^\ell$ is reparametrized before being applied to the convolution as $\frac{W^\ell}{\sqrt{2k_s +1}}$, where $k_s$ is the kernel size.

For our architecture the inputs from a layer come from either one or two different layers. In the case where the inputs come from only one layer, $\alpha$ the convexity constraint parameter is set to 1. When the inputs come from the sum of two layers we use $\alpha = 0.5$ as the value for both of them, which constraints our Lipschitz constant, this is softer than the convexity constraint from the original paper.

\section{Hyperparameters}

We train our networks using classical stochastic gradient descent with momentum ($0.9$), with batch size of $b=100$ images and using a L2-norm weight decay with a coefficient of $\lambda = 0.0005$. We do a 100 epoch training. Our learning rate starts at $0.1$. After half of the training~($50$ epochs) the learning rate decreases to $0.001$. 


We use the mean of the difference of smoothness between successive layers in our loss function. Therefore in our loss function we have: \begin{equation}\mathcal{L} = CategoricalCrossEntropy + \lambda WeightDecay + \gamma\Delta\end{equation} where  $\Delta = \frac{1}{d-1} \sum_{\ell=1}^d | \delta_\sigma^\ell |$. We perform experiments using various powers of the Laplacian $m = 1,2,3$, in which case the scaling coefficient $\gamma$ is put to the same power as the Laplacian.

We tested multiple parameters of $\beta$, the Parseval tightness parameter, $\gamma$ the weight for the smoothness difference cost and $m$ the power of the Laplacian. We found that the best values for this specific architecture, dataset and training scheme were: $\beta = 0.01, \gamma = 0.01, m = 2, k = b$.

\section{Depiction of the network}

\begin{figure}[ht]
  \label{Network}
  \caption{Depiction of the studied network}
    \begin{center}
  \begin{tikzpicture}[thick, scale=0.3]
    \tiny{\node (00) at (0,1) {};
    \node(0)[draw,rectangle] at (0,0) {Conv layer, $f$};
    \node(1)[draw,rectangle, imtatlantique] at (0,-2) {Conv layer, $f$};
    \node(2)[draw,rectangle, imtatlantique] at (0,-4) {Conv layer, $f$};
    \node(3)[draw,rectangle, red] at (0,-6) {Conv layer, $f$};

    \node(5)[draw,rectangle, red] at (0,-8) {Conv layer, $f$};
    \node(6)[draw,rectangle, green!50!black] at (0,-10) {Strided Conv layer, $2f$};
    \node(7)[draw,rectangle, green!50!black] at (0,-12) {Conv layer, $2f$};
    \node(8)[draw,rectangle, blue] at (0,-14) {Conv layer, $2f$};
    \node(9)[draw,rectangle, blue] at (0,-16) {Conv layer, $2f$};
    \node(10)[draw,rectangle, blue] at (0,-18) {Strided Conv layer, $4f$};
    \node(11) at (0,-18) {};
    \path[->, >=stealth']
    (00) edge (0)
    (0) edge (1)
    (1) edge (2)
    (2) edge (3)
    (3) edge (5)
    (5) edge (6)
    (6) edge (7)
    (7) edge (8)
    (8) edge (9)
    (9) edge (10)
    ;
    
    \draw[->,>=stealth', dashed]
    (0,-1) to[out=180, in=90] (-4, -3) to[out=-90, in=180] (0,-4.8);
    \draw[->,>=stealth', dashed]      
    (0,-5.2) to[out=180, in=90] (-4, -7) to[out=-90, in=180] (0,-8.8);
    \draw[->,>=stealth', dashed]      
    (0,-9.2) to[out=180, in=90] (-6, -11) to[out=-90, in=180] (0,-12.8);
    \draw[->,>=stealth', dashed]      
    (0,-13.2) to[out=180, in=90] (-8, -15) to[out=-90, in=180] (0,-16.8);
    \begin{scope}[xshift=10cm]
      \node (00) at (0,1) {};
      \node(0)[draw,rectangle, orange!50!black] at (0,0) {Conv layer, $4f$};
      \path[->,>=stealth']
      (10) edge[out = 0, in=180] (0);

      \node(1)[draw,rectangle, orange!50!black] at (0,-2) {Conv layer, $4f$};
      \node(2)[draw,rectangle, purple] at (0,-4) {Conv layer, $4f$};
      \node(3)[draw,rectangle, purple] at (0,-6) {Strided Conv layer, $8f$};

      \node(5)[draw,rectangle, black!80] at (0,-8) {Conv layer, $8f$};
      \node(6)[draw,rectangle, black!80] at (0,-10) {Conv layer, $8f$};
      \node(7)[draw,rectangle, red!50!green] at (0,-12) {Conv layer, $8f$};
      \node(8)[draw,rectangle] at (0,-14) {Global Avg pooling, $8f$};
      \node(9)[draw,rectangle] at (0,-16) {Linear+Softmax, $10$};

      \node(10) at (0,-18) {};
      
      \path[->, >=stealth']
      (0) edge (1)
      (1) edge (2)
      (2) edge (3)
      (3) edge (5)
      (5) edge (6)
      (6) edge (7)
      (7) edge (8)
      (8) edge (9);
      \path[->,>=stealth']
      (9) edge (10);
      \draw[->,>=stealth', dashed]
      (-10,-17.2) to (-5,-17.2) to (-5,-0.8) to (0,-0.8);
      \draw[->,>=stealth', dashed]
      (0,-1) to[out=0, in=90] (6, -3) to[in=0, out=270] (0,-4.8);
      \draw[->,>=stealth', dashed]      
      (0,-5.2) to[out=0, in=90] (6, -7) to[in=0, out=270] (0,-8.8);
      \draw[->,>=stealth', dashed]      
      (0,-9.2) to[out=0, in=90] (8, -11) to[in=0, out=270] (0,-12.8);


    \end{scope}}

  \end{tikzpicture}
  \end{center}
\end{figure}
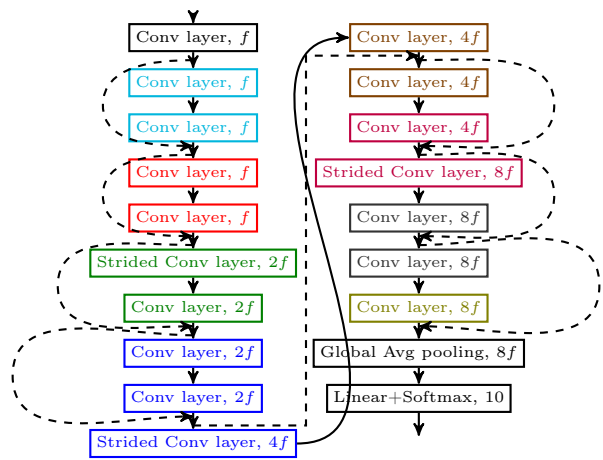

Figure~\ref{Network} depicts the network used on all experiments of this paper. $f=64$ is the filter size of the first layer of the network. Conv layers are 3x3 layers and are always preceded by batch normalization and relu (except for the first layer which receives just the input). The smoothness gaps are calculated after at each ReLU.

\end{document}